\def\year{2018}\relax
\documentclass[letterpaper]{article} 
\usepackage{aaai18}  
\usepackage{times}  
\usepackage{helvet}  
\usepackage{courier}  
\usepackage{url}  
\usepackage{graphicx}  
\usepackage{amsmath}
\usepackage{relsize}
\usepackage{hhline}
\usepackage{array}
\newcolumntype{M}[1]{>{\arraybackslash}m{#1}}
\usepackage{multirow}
\usepackage{booktabs}       
\usepackage{footnote}
\makesavenoteenv{tabular}
\makesavenoteenv{table}

\newcommand*{\Scale}[2][4]{\scalebox{#1}{$#2$}}%
\begin{document}
%
\title{Spatial As Deep: Spatial CNN for Traffic Scene Understanding}
\author{
Xingang Pan\textsuperscript{1}, 
Xiaohang Zhan\textsuperscript{1}, 
Jianping Shi\textsuperscript{2}, 
Ping Luo\textsuperscript{1}, 
Xiaogang Wang\textsuperscript{1}, \and
Xiaoou Tang\textsuperscript{1} \\
\textsuperscript{1}{The Chinese University of Hong Kong} \ \textsuperscript{2}{SenseTime Group Limited}
}
\maketitle
\begin{abstract}
Convolutional neural networks (CNNs) are usually built by stacking convolutional operations layer-by-layer. 
Although CNN has shown strong capability to extract semantics from raw pixels, its capacity to capture spatial relationships of pixels across rows and columns of an image is not fully explored. 
These relationships are important to learn semantic objects with strong shape priors but weak appearance coherences, such as traffic lanes, which are often occluded or not even painted on the road surface as shown in Fig.~\ref{cnn_fail} (a).
In this paper, we propose Spatial CNN (SCNN), which generalizes traditional deep layer-by-layer convolutions to slice-by-slice convolutions within feature maps, thus enabling message passings between pixels across rows and columns in a layer. 
Such SCNN is particular suitable for long continuous shape structure or large objects, with strong spatial relationship but less appearance clues, such as traffic lanes, poles, and wall. 
We apply SCNN on a newly released very challenging traffic lane detection dataset and Cityscapse dataset\footnote{Code is available at \textit{https://github.com/XingangPan/SCNN}}. 
The results show that SCNN could learn the spatial relationship for structure output and significantly improves the performance.
We show that SCNN outperforms the recurrent neural network (RNN) based ReNet and MRF+CNN (MRFNet) in the lane detection dataset by 8.7\% and 4.6\% respectively.
Moreover, our SCNN won the 1st place on the TuSimple Benchmark Lane Detection Challenge, with an accuracy of 96.53\%. 
\end{abstract}

\section{Introduction}
\begin{figure}[!t]
\centering
\includegraphics[width=8.5cm]{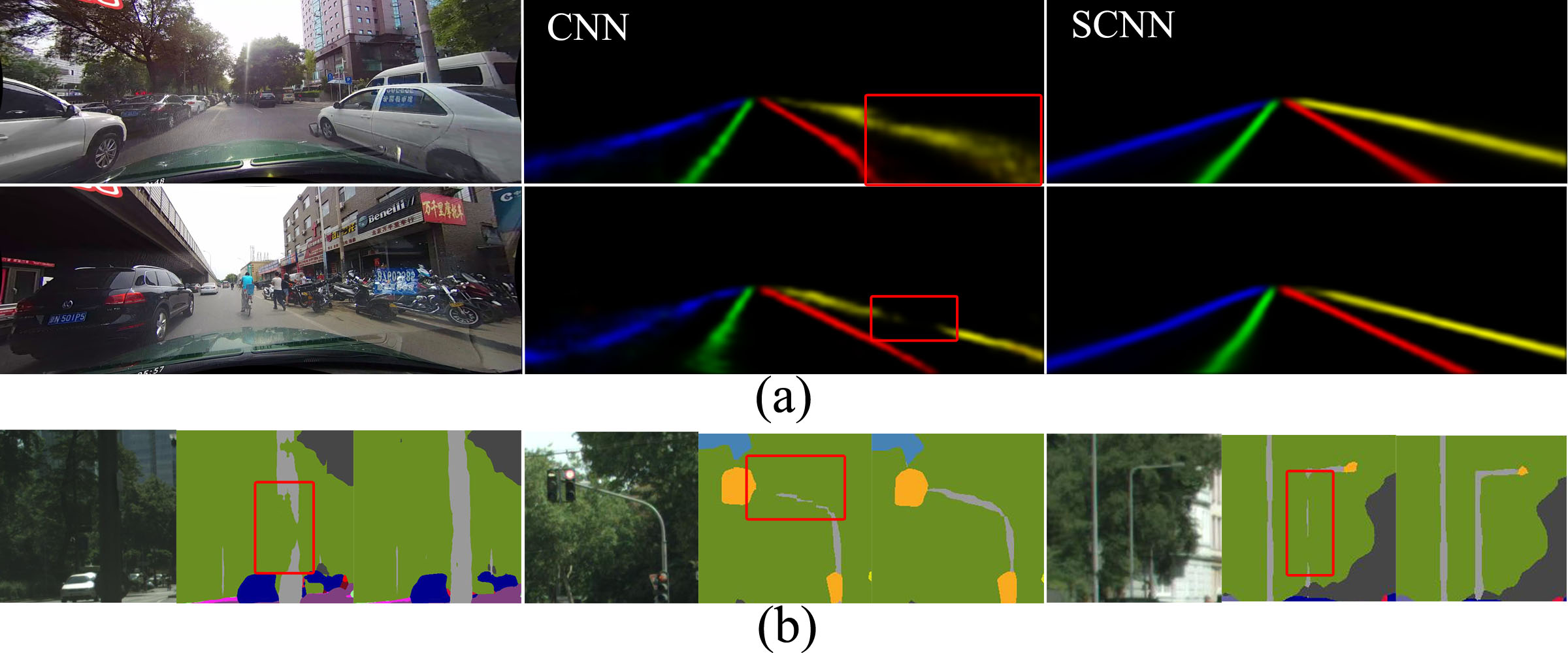}
\caption{\label{cnn_fail} Comparison between CNN and SCNN in (a) lane detection and (b) semantic segmentation. For each example, from left to right are: input image, output of CNN, output of SCNN. It can be seen that SCNN could better capture the long continuous shape prior of lane markings and poles and fix the disconnected parts in CNN. }
\end{figure}

In recent years, autonomous driving has received much attention in both academy and industry. 
One of the most challenging task of autonomous driving is traffic scene understanding, which comprises computer vision tasks like lane detection and semantic segmentation. 
Lane detection helps to guide vehicles and could be used in driving assistance system~\cite{urmson2008autonomous}, while semantic segmentation provides more detailed positions about surrounding objects like vehicles or pedestrians.
In real applications, however, these tasks could be very challenging considering the many harsh scenarios, including bad weather conditions, dim or dazzle light, etc.
Another challenge of traffic scene understanding is that in many cases, especially in lane detection, we need to tackle objects with strong structure prior but less appearance clues like lane markings and poles, which have long continuous shape and might be occluded. For instance, in the first example in Fig.~\ref{cnn_fail} (a), the car at the right side fully occludes the rightmost lane marking.

Although CNN based methods~\cite{krizhevsky2012imagenet,long2015fully} have pushed scene understanding to a new level thanks to the strong representation learning ability. 
It is still not performing well for objects having long structure region and could be occluded, such as the lane markings and poles shown in the red bounding boxes in Fig.~\ref{cnn_fail}.
However, humans can easily infer their positions and fill in the occluded part from the context, i.e., the viewable part. 

\begin{figure*}[!htb]
\centering
\includegraphics[width=15.5cm]{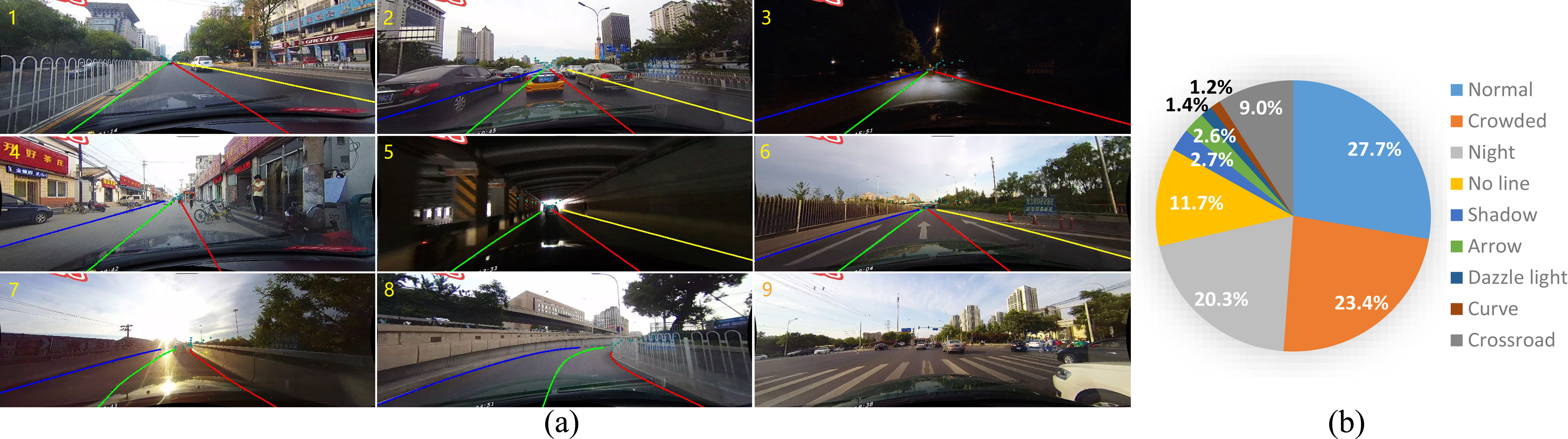}
\caption{\label{examples} (a) Dataset examples for different scenarios. (b) Proportion of each scenario.}
\end{figure*}

To address this issue, we propose Spatial CNN (SCNN), a generalization of deep convolutional neural networks to a rich spatial level. 
In a layer-by-layer CNN, a convolution layer receives input from the former layer, applies convolution operation and nonlinear activation, and sends result to the next layer.
This process is done sequentially.
Similarly, SCNN views rows or columns of feature maps as layers and applies convolution, nonlinear activation, and sum operations sequentially, which forms a deep neural network.
In this way information could be propagated between neurons in the same layer. 
It is particularly useful for structured object such as lanes, poles, or truck with occlusions, since the spatial information can be reinforced via inter layer propagation.
As shown in Fig.~\ref{cnn_fail}, in cases where CNN is discontinuous or is messy, SCNN could well preserve the smoothness and continuity of lane markings and poles.
In our experiment, SCNN significantly outperforms other RNN or MRF/CRF based methods, and also gives better results than the much deeper ResNet-101~\cite{he2016deep}.

\textbf{Related Work.}
For lane detection, most existing algorithms are based on hand-crafted low-level features~\cite{aly2008real,son2015real,jung2016efficient}, limiting there capability to deal with harsh conditions. 
Only \citeauthor{huval2015empirical} (\citeyear{huval2015empirical}) gave a primacy attempt adopting deep learning in lane detection but without a large and general dataset.
While for semantic segmentation, CNN based methods have become mainstream and achieved great success~\cite{long2015fully,chen2016deeplab}.

There have been some other attempts to utilize spatial information in neural networks.
\citeauthor{visin2015renet} (\citeyear{visin2015renet}) and \citeauthor{bell2016inside} (\citeyear{bell2016inside}) used recurrent neural networks to pass information along each row or column, thus in one RNN layer each pixel position could only receive information from the same row or column. 
\citeauthor{liang2016semantic} (\citeyear{liang2016semantic2,liang2016semantic}) proposed variants of LSTM to exploit contextual information in semantic object parsing, but such models are computationally expensive. 
Researchers also attempted to combine CNN with graphical models like MRF or CRF, in which message pass is realized by convolution with large kernels~\cite{liu2015semantic,tompson2014joint,chu2016crf}. 
There are three advantages of SCNN over these aforementioned methods: in SCNN, (1) the sequential message pass scheme is much more computational efficiency than traditional dense MRF/CRF, (2) the messages are propagated as residual, making SCNN easy to train, and (3) SCNN is flexible and could be applied to any level of a deep neural network.

\begin{figure*}[!t]
\centering
\includegraphics[width=13cm]{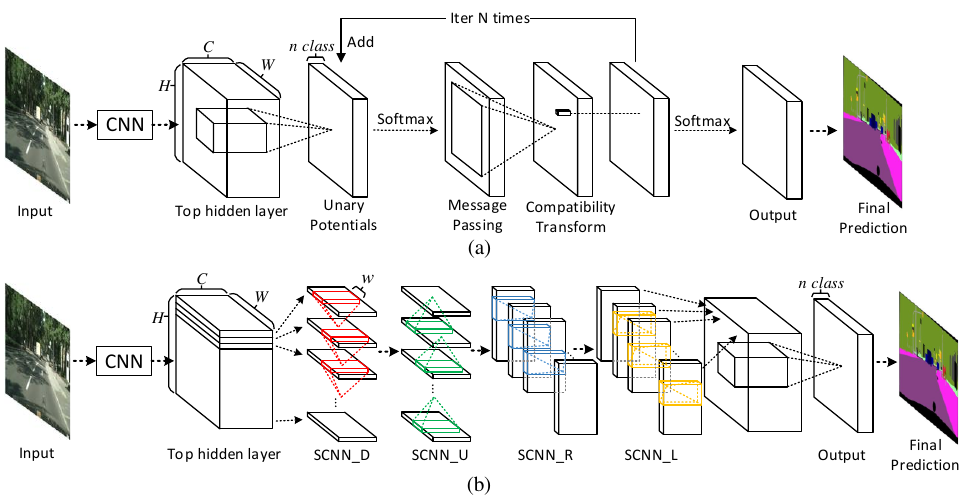}
\caption{\label{spatialCNN_model} (a) MRF/CRF based method. (b) Our implementation of Spatial CNN. MRF/CRF are theoretically applied to unary potentials whose channel number equals to the number of classes to be classified, while SCNN could be applied to the top hidden layers with richer information. }
\end{figure*}

\section{Spatial Convolutional Neural Network}
\subsection{Lane Detection Dataset}

In this paper, we present a large scale challenging dataset for traffic lane detection.
Despite the importance and difficulty of traffic lane detection, existing datasets are either too small or too simple, and a large public annotated benchmark is needed to compare different methods~\cite{bar2014recent}. 
KITTI~\cite{fritsch2013new} and CamVid~\cite{brostow2008segmentation} contains pixel level annotations for lane/lane markings, but have merely hundreds of images, too small for deep learning methods. 
Caltech Lanes Dataset~\cite{aly2008real} and the recently released TuSimple Benchmark Dataset~\cite{tusimple} consists of 1224 and 6408 images with annotated lane markings respectively, while the traffic is in a constrained scenario, which has light traffic and clear lane markings. 
Besides, none of these datasets annotates the lane markings that are occluded or are unseen because of abrasion, while such lane markings can be inferred by human and is of high value in real applications.

To collect data, we mounted cameras on six different vehicles driven by different drivers and recorded videos during driving in Beijing on different days. 
More than 55 hours of videos were collected and 133,235 frames were extracted, which is more than 20 times of TuSimple Dataset.
We have divided the dataset into 88880 for training set, 9675 for validation set, and 34680 for test set.
These images were undistorted using tools in \cite{scaramuzza2006flexible} and have a resolution of \(1640\times 590\). 
Fig.~\ref{examples} (a) shows some examples, which comprises urban, rural, and highway scenes.
As one of the largest and most crowded cities in the world, Beijing provides many challenging traffic scenarios for lane detection.
We divided the test set into normal and 8 challenging categories, which correspond to the 9 examples in Fig.~\ref{examples} (a).
Fig.~\ref{examples} (b) shows the proportion of each scenario. 
It can be seen that the 8 challenging scenarios account for most (72.3\%) of the dataset.

For each frame, we manually annotate the traffic lanes with cubic splines. 
As mentioned earlier, in many cases lane markings are occluded by vehicles or are unseen. 
In real applications it is important that lane detection algorithms could estimate lane positions from the context even in these challenging scenarios that occur frequently. 
Therefore, for these cases we still annotate the lanes according to the context, as shown in Fig.~\ref{examples} (a) (2)(4).
We also hope that our algorithm could distinguish barriers on the road, like the one in Fig.~\ref{examples} (a) (1). Thus the lanes on the other side of the barrier are not annotated. 
In this paper we focus our attention on the detection of four lane markings, which are paid most attention to in real applications.
Other lane markings are not annotated.

\subsection{Spatial CNN}

Traditional methods to model spatial relationship are based on Markov Random Fields (MRF) or Conditional Random Fields (CRF)~\cite{krahenbuhl2011efficient}.
Recent works~\cite{zheng2015conditional,liu2015semantic,chen2016deeplab} to combine them with CNN all follow the pipeline of Fig.~\ref{spatialCNN_model} (a), where the mean field algorithm can be implemented with neural networks.
Specifically, the procedure is (1) Normalize: the output of CNN is viewed as unary potentials and is normalized by the Softmax operation, (2) Message Passing, which could be realized by channel wise convolution with large kernels (for dense CRF, the kernel size would cover the whole image and the kernel weights are dependent on the input image), (3) Compatibility Transform, which could be implemented with a \(1\times 1\) convolution layer, and (4) Adding unary potentials. This process is iterated for N times to give the final output.

It can be seen that in the message passing process of traditional methods, each pixel receives information from all other pixels, which is very computational expensive and hard to be used in real time tasks as in autonomous driving.
For MRF, the large convolution kernel is hard to learn and usually requires careful initialization~\cite{tompson2014joint,liu2015semantic}.
Moreover, these methods are applied to the output of CNN, while the top hidden layer, which comprises richer information, might be a better place to model spatial relationship.

To address these issues, and to more efficiently learn the spatial relationship and the smooth, continuous prior of lane markings, or other structured object in the driving scenario, we propose Spatial CNN. Note that the 'spatial' here is not the same with that in 'spatial convolution', but denotes propagating spatial information via specially designed CNN structure.

As shown in the 'SCNN\_D' module of Fig.~\ref{spatialCNN_model} (b), considering a SCNN applied on a 3-D tensor of size \(C\times H\times W\), where \textit{C}, \textit{H}, and \textit{W} denote the number of channel, rows, and columns respectively.
The tensor would be splited into \textit{H} slices, and the first slice is then sent into a convolution layer with \textit{C} kernels of size \(C\times w\), where \textit{w} is the kernel width.
In a traditional CNN the output of a convolution layer is then fed into the next layer, while here the output is added to the next slice to provide a new slice.
The new slice is then sent to the next convolution layer and this process would continue until the last slice is updated.

Specifically, assume we have a 3-D kernel tensor \textbf{K} with element \(K_{i,j,k}\) denoting the weight between an element in channel \textit{i} of the last slice
and an element in channel \textit{j} of the current slice, with an offset of \textit{k} columes between two elements.
Also denote the element of input 3-D tensor \textbf{X} as \(X_{i,j,k}\), where \textit{i}, \textit{j}, and \textit{k} indicate indexes of channel, row, and column respectively. Then the forward computation of SCNN is:
\begin{equation}
	\begin{split}
	X_{i,j,k}' = 
	\begin{cases}
	X_{i,j,k}, \qquad\qquad\qquad\qquad\quad j = 1\\
	X_{i,j,k} + \mathlarger{f}\big(\sum\limits_{m} \sum\limits_{n} X_{\Scale[0.7]{m,j-1,k+n-1}}' \\
	\qquad\qquad\  \times K_{m,i,n}\big), \quad\  \Scale[0.9]{j = 2, 3, ..., H}
	\end{cases}
	\end{split}
\end{equation}
where \(f\) is a nonlinear activation function as ReLU.
The \(X\) with superscript \('\) denotes the element that has been updated. 
Note that the convolution kernel weights are shared across all slices, thus SCNN is a kind of recurrent neural network.
Also note that SCNN has directions.
In Fig.~\ref{spatialCNN_model} (b), the four 'SCNN' module with suffix 'D', 'U', 'R', 'L' denotes SCNN that is downward, upward, rightward, and leftward respectively.

\subsection{Analysis}

There are three main advantages of Spatial CNN over traditional methods, which are concluded as follows.

\begin{figure}[!t]
\centering
\includegraphics[width=5.6cm]{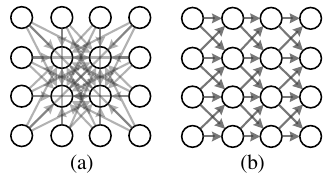}
\caption{\label{messagePass} Message passing directions in (a) dense MRF/CRF and (b) Spatial CNN (rightward). For (a), only message passing to the inner 4 pixels are shown for clearance. }
\end{figure}

(1) \textit{Computational efficiency.}
As show in Fig.~\ref{messagePass}, in dense MRF/CRF each pixel receives messages from all other pixels directly, which could have much redundancy, while in SCNN message passing is realized in a sequential propagation scheme. 
Specifically, assume a tensor with \textit{H} rows and \textit{W} columns, then in dense MRF/CRF, there is message pass between every two of the \(WH\) pixels.
For \(n_{iter}\) iterations, the number of message passing is \(n_{iter} W^2 H^2\).
In SCNN, each pixel only receive information from \textit{w} pixels, thus the number of message passing is \(n_{dir} W H w\), where \(n_{dir}\) and \textit{w} denotes the number of propagation directions in SCNN and the kernel width of SCNN respectively.
\(n_{iter}\) could range from 10 to 100, while in this paper \(n_{dir}\) is set to 4, corresponding to 4 directions, and \textit{w} is usually no larger than 10 (in the example in Fig.~\ref{messagePass} (b) \(w=3\)).
It can be seen that for images with hundreds of rows and columns, SCNN could save much computations, while each pixel still could receive messages from all other pixels with message propagation along 4 directions.

(2) \textit{Message as residual.}
In MRF/CRF, message passing is achieved via weighted sum of all pixels, which, according to the former paragraph, is computational expensive.
And recurrent neural network based methods might suffer from gradient descent~\cite{pascanu2013difficulty}, considering so many rows or columns.
However, deep residual learning~\cite{he2016deep} has shown its capability to easy the training of very deep neural networks.
Similarly, in our deep SCNN messages are propagated as residual, which is the output of ReLU in Eq.(1).
Such residual could also be viewed as a kind of modification to the original neuron.
As our experiments will show, such message pass scheme achieves better results than LSTM based methods.

(3) \textit{Flexibility}
Thanks to the computational efficiency of SCNN, it could be easily incorporated into any part of a CNN, rather than just output. 
Usually, the top hidden layer contains information that is both rich and of high semantics, thus is an ideal place to apply SCNN.
Typically, Fig.~\ref{spatialCNN_model} shows our implementation of SCNN on the LargeFOV~\cite{chen2016deeplab} model.
SCNNs on four spatial directions are added sequentially right after the top hidden layer ('fc7' layer) to introduce spatial message propagation.

\section{Experiment}

We evaluate SCNN on our lane detection dataset and Cityscapes~\cite{cordts2016cityscapes}. 
In both tasks, we train the models using standard SGD with batch size 12, base learning rate 0.01, momentum 0.9, and weight decay 0.0001. 
The learning rate policy is "poly" with power and iteration number set to 0.9 and 60K respectively.
Our models are modified based on the LargeFOV model in \cite{chen2016deeplab}. 
The initial weights of the first 13 convolution layers are copied from VGG16~\cite{Simonyan15} trained on ImageNet~\cite{deng2009imagenet}.
All experiments are implemented on the Torch7~\cite{collobert2011torch7} framework.

\subsection{Lane Detection}

\subsubsection{Lane detection model}

Unlike common object detection task that only requires bounding boxes, lane detection requires precise prediction of curves.
A natural idea is that the model should output probability maps (probmaps) of these curves, thus we generate pixel level targets to train the networks, like in semantic segmentation tasks. 
Instead of viewing different lane markings as one class and do clustering afterwards, we want the neural network to distinguish different lane markings on itself, which could be more robust. 
Thus these four lanes are viewed as different classes.
Moreover, the probmaps are then sent to a small network to give prediction on the existence of lane markings.

\begin{figure}[!t]
\centering
\includegraphics[width=8cm]{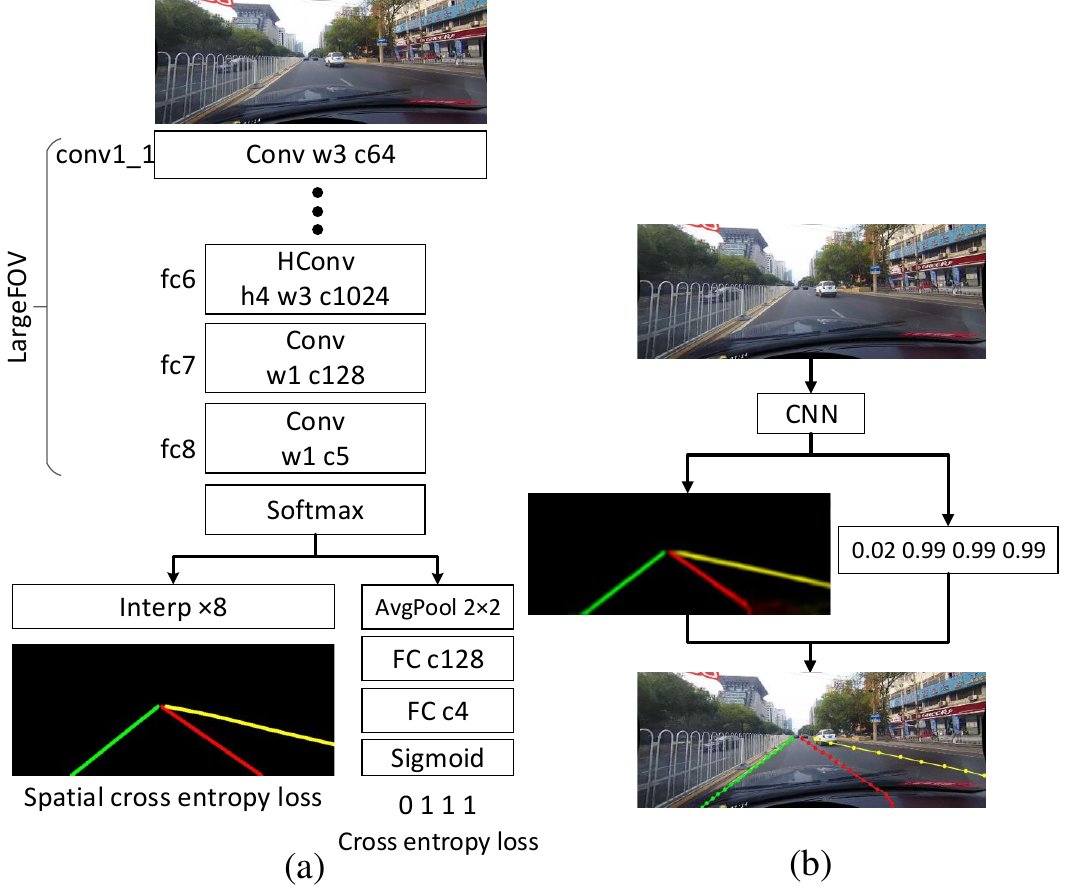}
\caption{\label{baseline}(a) Training model, (b) Lane prediction process. 'Conv','HConv', and 'FC' denotes convolution layer, atrous convolution layer~\cite{chen2016deeplab}, and fully connected layer respectively. 'c', 'w', and 'h' denotes number of output channels, kernel width, and 'rate' for atrous convolution.}
\end{figure}

During testing, we still need to go from probmaps to curves. As shown in Fig.\ref{baseline} (b), for each lane marking whose existence value is larger than 0.5, we search the corresponding probmap every 20 rows for the position with the highest response. These positions are then connected by cubic splines, which are the final predictions.

As shown in Fig.\ref{baseline} (a), the detailed differences between our baseline model and LargeFOV are: (1) the output channel number of the 'fc7' layer is set to 128, (2) the 'rate' for the atrous convolution layer of 'fc6' is set to 4, (3) batch normalization~\cite{ioffe2015batch} is added before each ReLU layer, (4) a small network is added to predict the existence of lane markings.
During training, the line width of the targets is set to 16 pixels, and the input and target images are rescaled to \(800\times 288\). Considering the imbalanced label between background and lane markings, the loss of background is multiplied by 0.4.

\subsubsection{Evaluation}

\begin{figure}[!t]
\centering
\includegraphics[width=8.5cm]{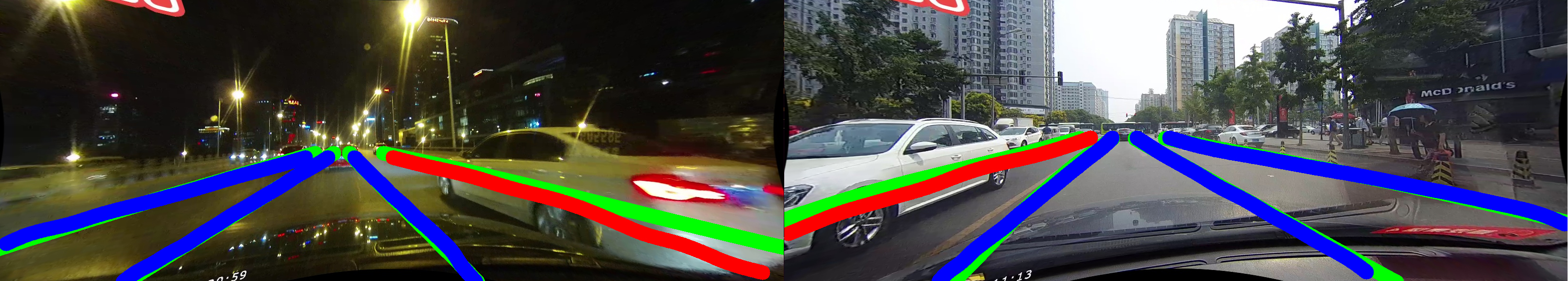}
\caption{\label{evaluation} Evaluation based on IoU. Green lines denote ground truth, while blue and red lines denote TP and FP respectively. }
\end{figure}

In order to judge whether a lane marking is successfully detected, we view lane markings as lines with widths equal to 30 pixel and calculate the intersection-over-union (IoU) between the ground truth and the prediction. 
Predictions whose IoUs are larger than certain threshold are viewed as true positives (TP), as shown in Fig.~\ref{evaluation}.
Here we consider 0.3 and 0.5 thresholds corresponding to loose and strict evaluations.
Then we employ \(\text{F-measure}=(1+\beta^2)\tfrac{\text{Precision Recall}}{\beta^2 \text{Precision} + \text{Recall}}\) as the final evaluation index, where \(\text{Precision}=\tfrac{TP}{TP + FP}\) and \(\text{Recall}=\tfrac{TP}{TP + FN}\).
Here \(\beta\) is set to 1, corresponding to harmonic mean (F1-measure).

\subsubsection{Ablation Study}

In section 2.2 we propose Spatial CNN to enable spatial message propagation.
To verify our method, we will make detailed ablation studies in this subsection. 
Our implementation of SCNN follows that shown in Fig.~\ref{spatialCNN_model}.

(1) \textit{Effectiveness of multidirectional SCNN.} 
Firstly, we investigate the effects of directions in SCNN.
We try SCNN that has different direction implementations, the results are shown in Table.~\ref{directions}.
Here the kernel width \textit{w} of SCNN is set to 5.
It can be seen that the performance increases as more directions are added.
To prove that the improvement does not result from more parameters but from the message passing scheme brought about by SCNN, we add an extra convolution layer with \(5\times 5\) kernel width after the top hidden layer of the baseline model and compare with our method.
From the results we can see that extra convolution layer could merely bring about little improvement, which verifies the effectiveness of SCNN.

\begin{table}[!h]
\small
  \caption{Experimental results on SCNN with different directional settings. F1 denotes F1-measure, and the value in the bracket denotes the IoU threshold. The suffix 'D', 'U', 'R', 'L' denote downward, upward, rightward, and leftward respectively. }
  \label{directions}
  \centering
  \resizebox{8cm}{0.8cm}{%
  \begin{tabular}{M{1.05cm}>{\centering\arraybackslash}M{0.83cm}>{\centering\arraybackslash}M{0.87cm}>{\centering\arraybackslash}M{0.85cm}
  >{\centering\arraybackslash}M{1cm}>{\centering\arraybackslash}M{1.05cm}}
    \toprule
    Models & {\fontsize{9}{7.2}\selectfont Baseline} & {\fontsize{8}{7.2}\selectfont ExtraConv} & {\fontsize{8}{7.2}\selectfont SCNN\_D} & {\fontsize{8}{7.2}\selectfont SCNN\_DU} & {\fontsize{8}{7.2}\selectfont SCNN\_DURL}  \\
    \midrule
    F1 (0.3)  & 77.7 & 77.6 & 79.5 & 79.9 & \textbf{80.2} \\
    F1 (0.5)  & 63.2 & 64.0 & 68.6 & 69.4 & \textbf{70.4} \\
    \bottomrule
  \end{tabular}
  }
\end{table}

(2) \textit{Effects of kernel width \textit{w}.}
We further try SCNN with different kernel width based on the "SCNN\_DURL" model, as shown in Table.~\ref{width}.
Here the kernel width denotes the number of pixels that a pixel could receive messages from, and the \(w=1\) case is similar to the methods in~\cite{visin2015renet,bell2016inside}.
The results show that larger \textit{w} is beneficial, and \(w=9\) gives a satisfactory result, which surpasses the baseline by a significant margin 8.4\% and 3.2\% corresponding to different IoU threshold.

\begin{table}[!h]
\small
  \caption{Experimental results on SCNN with different kernel widths. }
  \label{width}
  \centering
  \begin{tabular}{lcccccc}
    \toprule
    Kernel width \textit{w} & 1 & 3 & 5 & 7 & 9 & 11 \\
    \midrule
    F1 (0.3)  & 78.5 & 79.5 & 80.2 & 80.5 & \textbf{80.9} & 80.6 \\
    F1 (0.5)  & 66.3 & 68.9 & 70.4 & 71.2 & 71.6 & \textbf{71.7} \\
    \bottomrule
  \end{tabular}
\end{table}

(3) \textit{Spatial CNN on different positions.}
As mentioned earlier, SCNN could be added to any place of a neural network. 
Here we consider the SCNN\_DURL model applied on (1) output and (2) the top hidden layer, which correspond to Fig.~\ref{spatialCNN_model}.
The results in Table.~\ref{positions} indicate that the top hidden layer, which comprises richer information than the output, turns out to be a better position to apply SCNN.

\begin{table}[!h]
\small
  \caption{Experimental results on spatial CNN at different positions, with \(w=9\). }
  \label{positions}
  \centering
  \begin{tabular}{lcc}
    \toprule
    Position & Output  & Top hidden layer \\
    \midrule
    F1 (0.3)  & 79.9 & 80.9 \\
    F1 (0.5)  & 68.8 & 71.6 \\
    \bottomrule
  \end{tabular}
\end{table}

(4) \textit{Effectiveness of sequential propagation.}
In our SCNN, information is propagated in a sequential way, i.e., a slice does not pass information to the next slice until it has received information from former slices.
To verify the effectiveness of this scheme, we compare it with parallel propagation, i.e., each slice passes information to the next slice simultaneously before being updated. 
For this parallel case, the \('\) in the right part of Eq.(1) is removed.
As Table.~\ref{sequential} shows, the sequential message passing scheme outperforms the parallel scheme significantly.
This result indicates that in SCNN, a pixel does not merely affected by nearby pixels, but do receive information from further positions.

\begin{table}[!h]
\small
  \caption{Comparison between sequential and parallel message passing scheme, for SCNN\_DULR with \(w=9\). }
  \label{sequential}
  \centering
  \begin{tabular}{lcc}
    \toprule
    Message passing scheme & Parallel & Sequential \\
    \midrule
    F1 (0.3)  & 78.4 & 80.9 \\
    F1 (0.5)  & 65.2 & 71.6 \\
    \bottomrule
  \end{tabular}
\end{table}

\begin{table*}[!t]
\small
  \caption{Comparison with other methods, with IoU threshold=0.5. For crossroad, only FP are shown. }
  \label{others}
  \centering
  \resizebox{13.5cm}{2.1cm}{%
  \begin{tabular}{lccccccc}
    \toprule
    Category & Baseline & ReNet & DenseCRF & MRFNet & ResNet-50 & ResNet-101 & Baseline+SCNN \\
    \midrule
    Normal  & 83.1 & 83.3 & 81.3 & 86.3 & 87.4 & 90.2 & \textbf{90.6}\\
    Crowded  & 61.0 & 60.5 & 58.8 & 65.2 & 64.1 & 68.2 & \textbf{69.7}\\
    Night  & 56.9 & 56.3 & 54.2 &  61.3 & 60.6 & 65.9 & \textbf{66.1}\\
    No line  & 34.0 & 34.5 & 31.9 &  37.2 & 38.1 & 41.7 & \textbf{43.4}\\
    Shadow  & 54.7 & 55.0 & 56.3 & 59.3 & 60.7 & 64.6 & \textbf{66.9}\\
    Arrow  & 74.0 & 74.1 & 71.2 & 76.9 & 79.0 & 84.0 & \textbf{84.1}\\
    Dazzle light  & 49.9 & 48.2 & 46.2 & 53.7 & 54.1 & \textbf{59.8} & 58.5\\
    Curve  & 61.0 & 59.9 & 57.8 &  62.3 & 59.8 & \textbf{65.5} & 64.4\\
    Crossroad  & 2060 & 2296 & 2253 &  \textbf{1837} & 2505 & 2183 & 1990\\
    Total  & 63.2 & 62.9 & 61.0 & 67.0 & 66.7 & 70.8 & \textbf{71.6}\\
    \bottomrule
  \end{tabular}%
  }
\end{table*}

\begin{figure*}[!t]
\centering
\includegraphics[width=17.8cm]{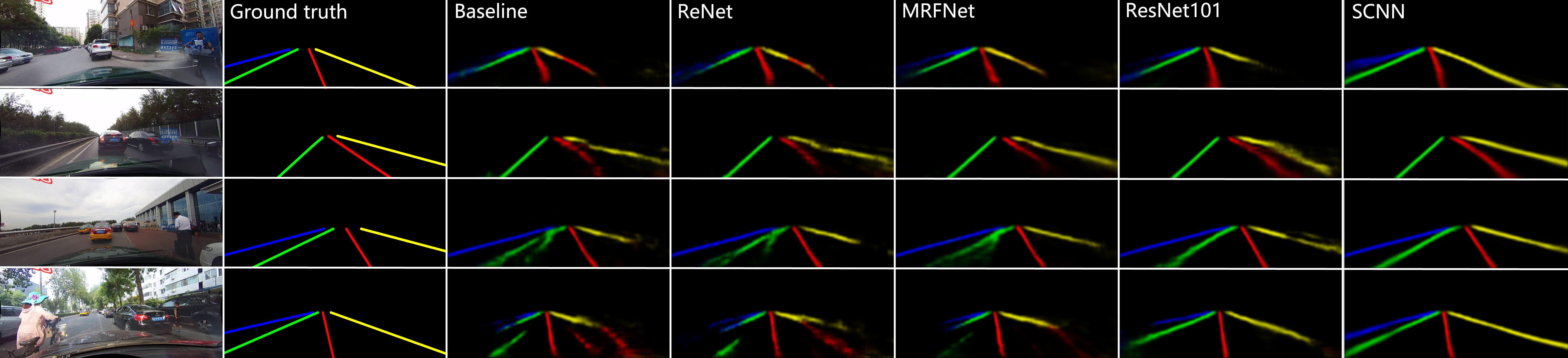}
\caption{\label{compare} Comparison between probmaps of baseline, ReNet, MRFNet, ResNet-101, and SCNN. }
\end{figure*}

(5) \textit{Comparison with state-of-the-art methods. }
To further verify the effectiveness of SCNN in lane detection, we compare it with several methods: the rnn based ReNet~\cite{visin2015renet}, the MRF based MRFNet, the DenseCRF~\cite{krahenbuhl2011efficient}, and the very deep residual network~\cite{he2016deep}.
For ReNet based on LSTM, we replace the "SCNN" layers in Fig.~\ref{spatialCNN_model} with two ReNet layers: one layer to pass horizontal information and the other to pass vertical information.
For DenseCRF, we use dense CRF as post-processing and employ 10 mean field iterations as in \cite{chen2016deeplab}.
For MRFNet, we use the implementation in Fig.~\ref{spatialCNN_model} (a), with iteration times and message passing kernel size set to 10 and 20 respectively.
The main difference of the MRF here with CRF is that the weights of message passing kernels are learned during training rather than depending on the image.
For ResNet, our implementation is the same with \cite{chen2016deeplab} except that we do not use the ASPP module.
For SCNN, we add SCNN\_DULR module to the baseline, and the kernel width \textit{w} is 9.
The test results on different scenarios are shown in Table~\ref{others}, and visualizations are given in Fig.~\ref{compare}. 

From the results, we can see that the performance of ReNet is not even comparable with SCNN\_DULR with \(w=1\), indicating the effectiveness of our residual message passing scheme.
Interestingly, DenseCRF leads to worse result here, because lane markings usually have less appearance clues so that dense CRF cannot distinguish lane markings and background.
In contrast, with kernel weights learned from data, MRFNet could to some extent smooth the results and improve performance, as Fig.~\ref{compare} shows, but are still not very satisfactory.
Furthermore, our method even outperform the much deeper ResNet-50 and ResNet-101.
Despite the over a hundred layers and the very large receptive field of ResNet-101, it still gives messy or discontinuous outputs in challenging cases, while our method, with only 16 convolution layers plus 4 SCNN layers, could preserve the smoothness and continuity of lane lines better.
This demonstrates the much stronger capability of SCNN to capture structure prior of objects over traditional CNN.

(6) \textit{Computational efficiency over other methods. }
In the \textbf{Analysis} section we give theoretical analysis on the computational efficiency of SCNN over dense CRF. 
To verify this, we compare their runtime experimentally. 
The results are shown in Table.~\ref{runtime}, where the runtime of the LSTM in ReNet is also given. 
Here the runtime does not include runtime of the backbone network.
For SCNN, we test both the practical case and the case with the same setting as dense CRF.
In the practical case, SCNN is applied on top hidden layer, thus the input has more channels but less hight and width.
In the fair comparison case, the input size is modified to be the same with that in dense CRF, and both methods are tested on CPU.
The results show that even in fair comparison case, SCNN is over 4 times faster than dense CRF, despite the efficient implementation of dense CRF in \cite{krahenbuhl2011efficient}.
This is because SCNN significantly reduces redundancy in message passing, as in Fig.~\ref{messagePass}.
Also, SCNN is more efficient than LSTM, whose gate mechanism requires more computation.

\begin{table}[!h]
\small
  \caption{Runtime of dense CRF, LSTM, and SCNN. The two SCNNs correspond to the one used in practice and the one whose input size is modified for fair comparison with dense CRF respectively. The kernel width \textit{w} of SCNN is 9. }
  \label{runtime}
  \centering
  \resizebox{8.4cm}{1.18cm}{%
  \begin{tabular}{l>{\centering\arraybackslash}M{1.6cm}>{\centering\arraybackslash}M{1.7cm}>{\centering\arraybackslash}M{1.7cm}>{\centering\arraybackslash}M{1.6cm}}
    \toprule
    Method & dense CRF & LSTM & \begin{tabular}[t]{@{}l@{}} SCNN\_DULR\\ {\fontsize{8}{7.2}\selectfont (in practice)}\end{tabular} & \begin{tabular}[t]{@{}l@{}} SCNN\_DULR\\ {\fontsize{8}{7.2}\selectfont (fair comparison)}\end{tabular} \\
    \midrule
    \begin{tabular}[t]{@{}l@{}}Input size\\ {\fontsize{7.5}{7.2}\selectfont \((C \times H \times W)\)}\end{tabular} & {\fontsize{8}{6.5}\selectfont \(5 \times 288 \times 800\)} & {\fontsize{8}{6.5}\selectfont \(128 \times 36 \times 100\)} & {\fontsize{8}{6.5}\selectfont \(128 \times 36 \times 100\)} & {\fontsize{8}{6.5}\selectfont \(5 \times 288 \times 800\)} \\
	Device      & CPU\footnote{\fontsize{7}{7.2}\selectfont Intel Core i7-4790K CPU} & GPU\footnote{\fontsize{7}{7.2}\selectfont GeForce GTX TITAN Black} & GPU & CPU \\    
    Runtime (ms) & 737 & 115 & 42 & 176 \\
    \bottomrule
  \end{tabular}
  }
\end{table}

\begin{table*}[!t]
\small
  \caption{Results on Cityscapes validation set. }
  \label{Cityscapes}
  \centering
  \resizebox{17.8cm}{1.18cm}{%
  \begin{tabular}{M{2.55cm}|>{\centering\arraybackslash}M{0.45cm}>{\centering\arraybackslash}M{0.6cm}>{\centering\arraybackslash}M{0.8cm}
  >{\centering\arraybackslash}M{0.6cm}>{\centering\arraybackslash}M{0.4cm}>{\centering\arraybackslash}M{0.5cm}>{\centering\arraybackslash}M{0.6cm}
  >{\centering\arraybackslash}M{0.6cm}>{\centering\arraybackslash}M{0.6cm}>{\centering\arraybackslash}M{0.75cm}>{\centering\arraybackslash}M{0.6cm}
  >{\centering\arraybackslash}M{0.45cm}>{\centering\arraybackslash}M{0.65cm}>{\centering\arraybackslash}M{0.9cm}>{\centering\arraybackslash}M{0.7cm}
  >{\centering\arraybackslash}M{0.4cm}>{\centering\arraybackslash}M{0.5cm}>{\centering\arraybackslash}M{0.55cm}>{\centering\arraybackslash}M{0.7cm}
  |>{\centering\arraybackslash}M{0.55cm}}
    \toprule
    Method & road & terrain & building & wall & car & pole & \begin{tabular}[t]{@{}l@{}}traffic\\light\end{tabular} & \begin{tabular}[t]{@{}l@{}}traffic\\sign\end{tabular} & fence & sidewalk & sky & rider & person & vegetation & truck & bus & train & motor & bicycle & \textbf{mIoU} \\
    \midrule
    LargeFOV & 97.0 & 59.2 & 89.9 & 42.2 & 92.3 & 52.9 & 62.3 & 71.1 & 52.2 & \textbf{78.8} & 92.2 & 52.1 & 75.9 & 91.0 & 48.8 & 70.2 & 37.6 & 54.6 & 72.3 & 68.0 \\
    LargeFOV+SCNN & 97.0 & \textbf{59.8} & \textbf{90.3} & \textbf{45.7} & \textbf{92.5} & \textbf{55.2} & 62.3 & \textbf{71.7} & \textbf{52.5} & 78.1 & \textbf{92.6} & \textbf{53.2} & \textbf{76.4} & \textbf{91.1} & \textbf{55.6} & \textbf{71.2} & \textbf{41.7} & \textbf{56.2} & 72.3 & \textbf{69.2} \\
    \midrule
    ResNet-101 & 98.3 & 64.2 & 92.4 & 44.5 & \textbf{94.9} & 66.0 & \textbf{74.5} & \textbf{82.1} & 59.9 & 86.0 & 94.7 & 65.5 & \textbf{84.1} & 92.7 & 57.3 & 81.1 & 54.0 & 64.5 & 80.0 & 75.6 \\
    ResNet-101+SCNN & 98.3 & \textbf{65.4} & \textbf{92.6} & \textbf{46.7} & 94.8 & \textbf{66.1} & 74.3 & 81.5 & \textbf{61.2} & \textbf{86.1} & 94.7 & 65.5 & 84.0 & 92.7 & \textbf{57.7} & \textbf{82.0} & \textbf{59.9} & \textbf{67.0} & \textbf{80.1} & \textbf{76.4} \\    
    \bottomrule
  \end{tabular}%
  }
\end{table*}

\begin{figure*}[!t]
\centering
\includegraphics[width=14cm]{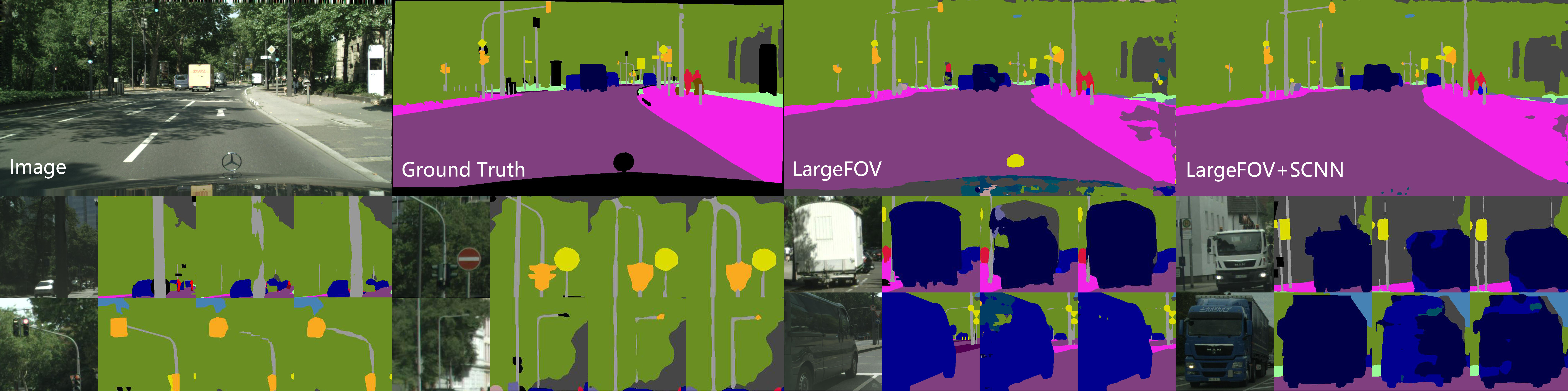}
\caption{\label{cityscapes} Visual improvements on Cityscapes validation set. For each example, from left to right are: input image, ground truth, result of LargeFOV, result of LargeFOV+SCNN.}
\end{figure*}

\subsection{Semantic Segmentation on Cityscapes}

To demonstrate the generality of our method, we also evaluate Spatial CNN on Cityscapes~\cite{cordts2016cityscapes}.
Cityscapes is a standard benchmark dataset for semantic segmentation on urban traffic scenes.
It contains 5000 fine annotated images, including 2975 for training, 500 for validation and 1525 for testing.
19 categories are defined including both stuff and objects.
We use two classic models, the LargeFOV and ResNet-101 in DeepLab~\cite{chen2016deeplab} as the baselines.
Batch normalization layers~\cite{ioffe2015batch} are added to LargeFOV to enable faster convergence.
For both models, the channel numbers of the top hidden layers are modified to 128 to make them compacter.

We add SCNN to the baseline models in the same way as in lane detection.
The comparisons between baselines and those combined with the SCNN\_DURL models with kernel width \(w=9\) are shown in Table~\ref{Cityscapes}.
It can be seen that SCNN could also improve semantic segmentation results.
With SCNNs added, the IoUs for all classes are at least comparable to the baselines, while the "wall", "pole", "truck", "bus", "train", and "motor" categories achieve significant improve.
This is because for long shaped objects like train and pole, SCNN could capture its continuous structure and connect the disconnected part, as shown in Fig.~\ref{cityscapes}.
And for wall, truck, and bus which could occupy large image area, the diffusion effect of SCNN could correct the part that are misclassified according to the context.
This shows that SCNN is useful not only for long thin structure, but also for large objects which require global information to be classified correctly.
There is another interesting phenomenon that the head of the vehicle at the bottom of the images, whose label is ignored during training, is in a mess in LargeFOV while with SCNN added it is classified as road.
This is also due to the diffusion effects of SCNN, which passes the information of road to the vehicle head area.

\begin{table}[!h]
\small
  \caption{Comparison between our SCNN and other MRF/CRF based methods on Cityscapes test set. }
  \label{CityscapesTest}
  \centering
  \resizebox{8cm}{0.75cm}{
  \begin{tabular}{M{1.1cm}>{\centering\arraybackslash}M{1.8cm}>{\centering\arraybackslash}M{1.8cm}>{\centering\arraybackslash}M{1.8cm}}
    \toprule
    Method & \begin{tabular}[t]{@{}c@{}}LargeFOV\\ {\fontsize{8}{7.2}\selectfont \cite{chen2016deeplab}}\end{tabular} & \begin{tabular}[t]{@{}c@{}}DPN\\ {\fontsize{9}{7.2}\selectfont \cite{liu2015semantic}}\end{tabular} & Ours \\
    \midrule
    mIoU  & 63.1 & 66.8 & 68.2 \\
    \bottomrule
  \end{tabular}
  }
\end{table}

To compare our method with other MRF/CRF based methods, we evaluate LargeFOV+SCNN on Cityscapes test set, and compare with methods that also use VGG16~\cite{Simonyan15} as the backbone network.
The results are shown in Table~\ref{CityscapesTest}.
Here LargeFOV, DPN, and our method use dense CRF, dense MRF, and SCNN respectively, and share nearly the same base CNN part.
The results show that our method achieves significant better performance.

\section{Conclusion}

In this paper, we propose Spatial CNN, a CNN-like scheme to achieve effective information propagation in the spatial level.
SCNN could be easily incorporated into deep neural networks and trained end-to-end.
It is evaluated at two tasks in traffic scene understanding: lane detection and semantic segmentation.
The results show that SCNN could effectively preserve the continuity of long thin structure, while in semantic segmentation its diffusion effects is also proved to be beneficial for large objects.
Specifically, by introducing SCNN into the LargeFOV model, our 20-layer network outperforms ReNet, MRF, and the very deep ResNet-101 in lane detection.
Last but not least, we believe that the large challenging lane detection dataset we presented would push forward researches on autonomous driving.

\section{Acknowledgments}

This work is supported by SenseTime Group Limited.
We would like to thank Jun Li, and Xudong Cao for helpful work in building the lane detection dataset.

{
\small
\bibliographystyle{aaai}
\bibliography{bib.bib}
}

\end{document}